\documentclass[10pt,twocolumn,letterpaper]{article}

\usepackage{cvpr}              

\usepackage{multirow}    
\usepackage{booktabs}    
\usepackage{tabularx}    
\usepackage{graphicx}    
\usepackage{CJKutf8}
\usepackage[accsupp]{axessibility}

\definecolor{cvprblue}{rgb}{0.21,0.49,0.74}
\usepackage[pagebackref,breaklinks,colorlinks,allcolors=cvprblue]{hyperref}

\def\paperID{18} 
\def\confName{CVPR}
\def\confYear{2026}

\title{Bias Redistribution in Visual Machine Unlearning: Does Forgetting \\ One Group Harm Another?}

\author{Yunusa Haruna\\
NewraLab, Suzhou, China\\
{\tt\small yunusa2k2@newralab.org}
\and
Adamu Lawan\\
Beihang University \\ Beijing GoerTek Alpha Lab\\ NewraLab, Suzhou, China\\
{\tt\small alawan@buaa.edu.cn}
\and
Ibrahim Haruna Abdulhamid\\
NewraLab, Suzhou, China\\
{\tt\small ibrahimharuna232@gmail.com}
\and
Hamza Mohammed Dauda\\
Skyline University Nigeria\\
{\tt\small hamzalo@yahoo.com}
\and
Jiaquan Zhang\\
UESTC, Chengdu, China\\
{\tt\small jiaquanzhang2005@gmail.com}
\and
Chaoning Zhang\\
UESTC, Chengdu, China\\
{\tt\small chaoningzhang1990@gmail.com}
\and
\\
Shamsuddeen Hassan Muhammad\\
Imperial College London\\
{\tt\small s.muhammad@imperial.ac.uk}
}
\begin{document}
\maketitle

\begin{abstract} 
Machine unlearning enables models to selectively forget training data, driven by privacy regulations such as GDPR and CCPA. However, its fairness implications remain underexplored: when a model forgets a demographic group, does it neutralize that concept or redistribute it to correlated groups, potentially amplifying bias? We investigate this bias redistribution phenomenon on CelebA using CLIP models (ViT-B/32, ViT-L/14, ViT-B/16) under a zero-shot classification setting across intersectional groups defined by age and gender. We evaluate three unlearning methods, Prompt Erasure, Prompt Reweighting, and Refusal Vector using per-group accuracy shifts, demographic parity gaps, and a redistribution score. Our results show that unlearning does not eliminate bias but redistributes it primarily along gender rather than age boundaries. In particular, removing the dominant Young Female group consistently transfers performance to Old Female across all model scales, revealing a gender-dominant structure in CLIP’s embedding space. While the Refusal Vector method reduces redistribution, it fails to achieve complete forgetting and significantly degrades retained performance. These findings highlight a fundamental limitation of current unlearning methods: without accounting for embedding geometry, they risk amplifying bias in retained groups.
\begin{figure}[t]
\centering
    \includegraphics[width=\linewidth]{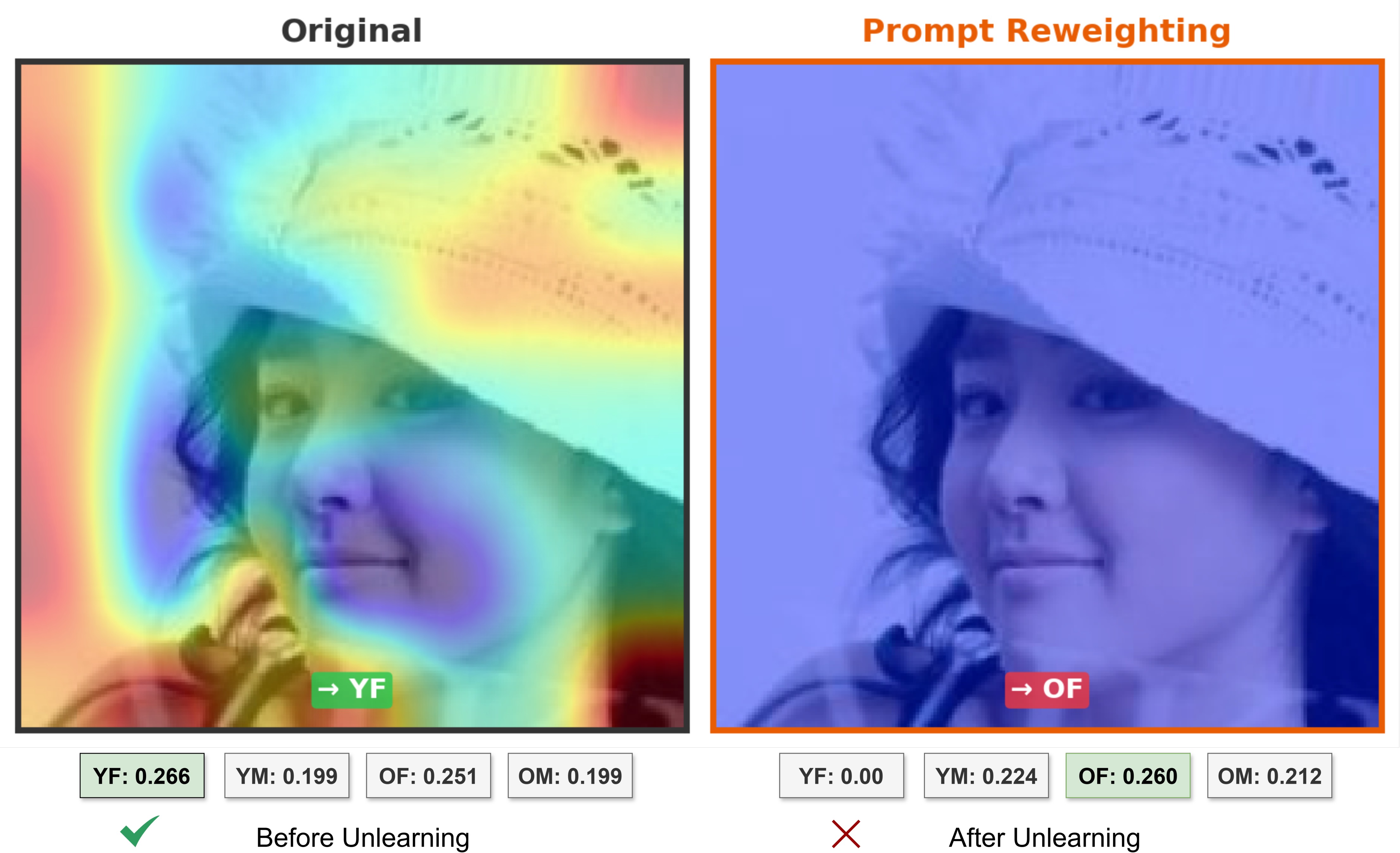}
    \caption{Bias redistribution in visual machine unlearning. \textbf{Left:} A Young Female face is correctly classified before unlearning, with strong patch-level similarity to the \textit{``young woman''} text prompt. \textbf{Right:} After applying Prompt Reweighting to forget the Young Female group, the same face is misclassified as \textit{Old Female}  the heatmap signal disappear, and probability mass has redistributed to the nearest retained group in CLIP's embedding space. This occurs because same-gender pairs are geometrically closer ($\cos(\text{YF},\text{OF}){=}0.945$) than same-age pairs ($\cos(\text{YF},\text{YM}){=}0.885$), directing redistribution along gender rather than age boundaries. See \S\ref{sec:analysis}. }
    \label{fig:motivation}
\end{figure}
\end{abstract}
\newpage
\section{Introduction}
\label{sec:intro} 
The right to be forgotten~\cite{voigt2017gdpr} has motivated a growing body of work on machine unlearning~\cite{cao2015towards,bourtoule2021machine}, which seeks to remove the influence of specific training samples from a deployed model without full retraining. In computer vision, unlearning has been applied to face recognition
privacy~\cite{gandikota2023erasing}, generative model concept
erasure~\cite{gandikota2023erasing}, and fairness-aware
adaptation~\cite{kearns2018preventing}.
Despite this progress, a critical question has gone largely unanswered:
\textbf{what happens to demographic fairness when a model unlearns a
protected group?}

Intuition suggests that forgetting group $\mathcal{D}_f$ should reduce
the model's reliance on features correlated with that group.
Yet in practice, those features may be entangled with correlated groups
in the model's representation space~\cite{buolamwini2018gender}.
Selectively suppressing $\mathcal{D}_f$ may cause the model to
over-predict a correlated group on ambiguous inputs  a phenomenon
we term \emph{bias redistribution}.
As illustrated in Figure~\ref{fig:motivation}, we find that making
CLIP~\cite{radford2021learning} forget \emph{Young Female} transfers
over 60 percentage points of classification accuracy onto
\emph{Old Female} rather than onto \emph{Young Male}  revealing
that CLIP's embedding space organizes faces along gender boundaries
more strongly than age boundaries, with same-gender pairs nearly
six points more similar than same-age pairs in cosine similarity.
This redistribution occurs not because the model learns anything new,
but because the geometry of the pretrained representation space
determines where probability mass flows when a class is suppressed.

We study this phenomenon across three CLIP model
variants~\cite{radford2021learning}  ViT-B/32, ViT-L/14, and
ViT-B/16  under a zero-shot classification setting, which allows
us to isolate the geometric properties of the embedding space from
training dynamics.
We apply three zero-shot unlearning strategies that differ in where
and how they modify the classifier: at the text embedding level
(Prompt Erasure~\cite{graves2021amnesiac} and Prompt
Reweighting~\cite{warnecke2021machine}) and at the image embedding
level (Refusal Vector~\cite{belrose2023leace}), to forget the Young
Female demographic group on CelebA~\cite{liu2015deep} and measure
the resulting shifts in per-group accuracy, demographic parity, and
a novel \emph{redistribution score}.
We further show that projection-based unlearning cannot achieve
perfect forgetting when the forget and retain distributions are
geometrically entangled  a fundamental impossibility rooted in
the collinearity of pretrained embeddings. The contributions of this work are:
\begin{itemize}
    \item We formally define bias redistribution in the context of
    machine unlearning and introduce the redistribution score to
    quantify it (\S\ref{sec:formulation}).

    \item We provide the first systematic empirical study of bias
    redistribution across three zero-shot unlearning methods on the
    CelebA benchmark, showing that redistribution consistently follows
    gender rather than age boundaries across three CLIP model scales,
    explained by a gender-dominant geometry in the embedding space
    ($\cos(\text{YF}, \text{OF}) = 0.945 > \cos(\text{YF},
    \text{YM}) = 0.885$) (\S\ref{sec:experiments}).

    \item We prove geometrically that projection-based unlearning
    cannot achieve perfect forgetting when forget and retain mean
    embeddings are nearly collinear
    ($\cos(\boldsymbol{\mu}_f, \boldsymbol{\mu}_r) = 0.929$),
    establishing a lower bound on achievable forget accuracy for
    any linear erasure method (\S\ref{sec:analysis}).

    \item We characterize the forget--fairness tradeoff via a
    continuous strength sweep of the Refusal Vector method,
    revealing a non-monotonic Pareto frontier where over-projection
    paradoxically restores the original geometry, and demonstrate
    that no zero-shot method simultaneously achieves low forget
    accuracy and low redistribution (\S\ref{sec:analysis}).
\end{itemize}
\section{Related Work}
\label{sec:related}

\noindent\textbf{Machine Unlearning.}
Cao and Yang~\cite{cao2015towards} introduced the formal unlearning problem, framing it as the removal of a training sample's influence
without full retraining. Bourtoule~\etal~\cite{bourtoule2021machine} proposed SISA training for exact unlearning via data sharding. Approximate unlearning methods include Gradient Ascent~\cite{graves2021amnesiac}, fine-tuning on the retain set~\cite{warnecke2021machine}, and NegGrad+~\cite{kurmanji2023towards}, which combines gradient ascent on the forget set with gradient descent on the retain set. Recent work has further expanded the approximate unlearning toolbox with retrain-free dampening methods such as Selective Synaptic Dampening (SSD)~\cite{foster2024fast}, saliency-guided unlearning such as SalUn~\cite{fan2023salun}, and broader surveys and benchmarks that systematize the design space and evaluation of unlearning algorithms~\cite{cadet2025deep,liu2024survey}. For concept-level unlearning in vision, Gandikota~\etal~\cite{gandikota2023erasing} erase visual concepts from diffusion models by fine-tuning text embeddings, a method closely related to our Prompt Erasure baseline. Related representation-space removal methods include iterative nullspace projection~\cite{ravfogel2020null}. In particular, the theoretical analysis of Belrose~\etal~\cite{belrose2023leace} establishes that perfect linear erasure requires favorable geometry between forget and retain directions, a condition we demonstrate is not satisfied in CLIP's pretrained embedding space, where forget and retain mean embeddings exhibit high cosine similarity ($\cos(\boldsymbol{\mu}_f, \boldsymbol{\mu}_r) = 0.929$), rendering perfect forgetting geometrically unachievable.

\noindent\textbf{Fairness in Vision Models.}
Demographic bias in face recognition is well
documented~\cite{buolamwini2018gender}, with error rates varying
substantially across intersectional subgroups defined by gender and
skin tone. Intersectional bias, where multiple protected attributes compound
unfairness~\cite{kearns2018preventing}, is central to our setting,
as we evaluate across four demographic slices defined by the joint
\textit{Young} $\times$ \textit{Male} attributes in
CelebA~\cite{liu2015deep}. Related benchmark efforts such as
FairFace~\cite{karkkainen2021fairface} and FACET~\cite{gustafson2023facet}
further show that demographic disparities persist across age, gender,
and race in modern vision systems. More broadly, recent vision research
has also explored settings beyond standard face benchmarks, including
automatically constructed real-world datasets, restoration-oriented
visual pipelines, and alternative visual architectures, which may
provide complementary context for understanding model behavior in
broader visual environments~\cite{sun2025datasetagent,wang2026sd,wang2025spiking,zhang2026learning}.
Recent work has shown that large
vision-language models such as CLIP~\cite{radford2021learning,wang2022fairclip} inherit
and amplify societal biases present in their pretraining data, and that
these biases can be probed and mitigated at the representation level
through methods such as DeAR~\cite{seth2023dear} and
FairCLIP~\cite{luo2024fairclip}. These results suggest that
demographic concepts in CLIP are encoded in a geometry that reflects
social correlations rather than independent attribute axes, making CLIP
a natural testbed for studying how unlearning interacts with pretrained
demographic representations.

\noindent\textbf{Unlearning and Fairness.}
While differential privacy provides formal guarantees against
membership inference~\cite{voigt2017gdpr,dwork2006calibrating}, it does not address
how forgetting one group affects predictions on retained groups.
Fair machine learning has studied how to \emph{prevent} bias at
training time~\cite{kearns2018preventing}, but the downstream fairness
consequences of post-hoc unlearning remain largely underexplored.
Existing unlearning benchmarks evaluate forgetting quality and
utility preservation~\cite{bourtoule2021machine, kurmanji2023towards}
but do not measure redistribution of classification mass onto
correlated retained groups. Recent work has begun to examine fairness-aware
unlearning under group-skewed forget sets, for example through
group-robust machine unlearning~\cite{de2025group}.
However,  no prior work directly studies bias redistribution
as a consequence of machine unlearning, that is, the transfer of
classification mass from a forgotten group onto correlated retained
groups, nor the geometric conditions under which such redistribution
is provably unavoidable. This gap motivates our study.

\section{Problem Formulation}
\label{sec:formulation}

\noindent\textbf{Setup.}
Let $\mathcal{M}$ be a zero-shot classifier built on a pretrained
vision-language model.
Given a set of $K$ demographic text prompts $\{p_1, \ldots, p_K\}$,
the classifier encodes each prompt into a normalized text embedding
$\mathbf{w}_k = \text{enc}_\text{text}(p_k) /
\|\text{enc}_\text{text}(p_k)\|$, forming a classifier weight matrix
$\mathbf{W} \in \mathbb{R}^{K \times d}$, where $d$ is the
model-specific embedding dimension.
For an image $x$, the predicted class is:
\begin{equation}
    \hat{y}(x) = \operatorname*{argmax}_k \;
    \frac{\text{enc}_\text{img}(x)}{\|\text{enc}_\text{img}(x)\|}
    \cdot \mathbf{w}_k^\top.
    \label{eq:zeroshot}
\end{equation}

An unlearning method $\mathcal{U}$ produces a modified classifier
$\mathcal{M}_\mathcal{U}$ by manipulating $\mathbf{W}$, the image
embeddings, or both, without any gradient-based retraining.
We instantiate this framework on three CLIP model
variants~\cite{radford2021learning}: ViT-B/32 ($d{=}512$),
ViT-L/14 ($d{=}768$), and ViT-B/16 ($d{=}512$).

\noindent\textbf{Demographic Groups.}
We define $K{=}4$ intersectional demographic groups over the
\textit{Young} $\times$ \textit{Male} attribute axes of
CelebA~\cite{liu2015deep}: Young Female~(YF), Young Male~(YM),
Old Female~(OF), and Old Male~(OM).
The forget target is $G_t = \text{YF}$, the group on which all three
baseline models achieve their highest accuracy ($\geq 97\%$),
indicating strong representational dominance across model scales.
The retain set comprises the remaining three groups:
$\mathcal{D}_r = \{G_k : k \neq t\}$.

\noindent\textbf{Bias Redistribution.}
We define \emph{bias redistribution} as a statistically meaningful
change in per-group accuracy for retained groups $G_k \neq G_t$
after applying $\mathcal{U}$.
Formally, redistribution occurs for group $G_k$ if:
\begin{equation}
    |\,\text{Acc}(\mathcal{M}_\mathcal{U},\, G_k) -
      \text{Acc}(\mathcal{M},\, G_k)\,| > \epsilon,
    \quad k \neq t,
    \label{eq:redistribution}
\end{equation}
where we set $\epsilon = 2\%$ as the significance threshold.
In our experiments, all three methods exceed this threshold on at
least one retained group by a wide margin (up to $71.19$ percentage
points), confirming systematic redistribution rather than noise.

\noindent\textbf{Metrics.}
We evaluate unlearning quality and fairness jointly using:
\begin{itemize}
    \item Forget Accuracy (FA): accuracy of
    $\mathcal{M}_\mathcal{U}$ on $G_t$ after unlearning
    (lower $=$ better forgetting).

    \item Retain Accuracy (RA): mean accuracy across all
    retained groups $G_k,\, k \neq t$
    (higher $=$ better utility preservation).

    \item Per-Group Accuracy Shift ($\Delta\text{Acc}_k$):
    signed accuracy change for each retained group $G_k$ relative
    to $\mathcal{M}$, revealing the direction and magnitude of
    redistribution.

    \item Demographic Parity Gap (DP) \cite{dwork2012fairness}:
    $\max_{i,j} |\hat{P}(\hat{y}{=}G_i \mid x \in G_i) -
    \hat{P}(\hat{y}{=}G_j \mid x \in G_j)|$,
    measuring the spread of per-group classification rates before and after unlearning.

    \item Redistribution Score (RS): mean absolute
    per-group accuracy shift across all retained groups:
    \begin{equation}
        \text{RS} = \frac{1}{K-1}
        \sum_{k \neq t} |\Delta\text{Acc}_k|.
        \label{eq:rs}
    \end{equation}
    
    A higher RS indicates more severe bias redistribution.
\end{itemize}
\section{Unlearning Methods}
\label{sec:methods}

We evaluate three zero-shot unlearning methods that differ in where and how they modify the classifier: at the text embedding level (Prompt Erasure and Prompt Reweighting) and at the image embedding level (Refusal Vector). Each method operates on the weight matrix
$\mathbf{W} \in \mathbb{R}^{K \times d}$ defined in
\S\ref{sec:formulation}, the image embedding space, or both,
without any gradient-based retraining.
All three methods are model-agnostic and are applied identically
across the three CLIP variants evaluated in this work.

\noindent\textbf{Prompt Erasure (PE).}
PE zeros out forget group's text embedding in $\mathbf{W}$,
prevents classifier from ever predicting $G_t$:
\begin{equation}
    \mathbf{w}_t \leftarrow \mathbf{0}.
    \label{eq:pe}
\end{equation}

This is the zero-shot analogue of Gradient
Ascent~\cite{graves2021amnesiac} maximally aggressive,
guaranteeing $\text{FA} = 0\%$, but forcing all probability mass
previously assigned to $G_t$ to redistribute across retained groups
according to their cosine similarity to the input image.

\noindent\textbf{Prompt Reweighting (PR).}
PR redistributes the forget embedding's mass to retained groups
proportionally to their cosine similarity to $\mathbf{w}_t$,
rather than discarding it entirely:
\begin{equation}
    \mathbf{w}_k \leftarrow
    \mathrm{normalize}\!\left(
        \mathbf{w}_k + \alpha\, s_k\, \mathbf{w}_t
    \right), \quad k \neq t,
    \label{eq:pr}
\end{equation}
where $s_k = \mathrm{softmax}_k\!\left(\cos(\mathbf{w}_t,
\mathbf{w}_k) / \tau\right)$ with temperature $\tau{=}0.07$,
and $\alpha{=}1.0$.
The forget embedding is then zeroed:
$\mathbf{w}_t \leftarrow \mathbf{0}$.
This is the zero-shot analogue of Retain-Set
Fine-tuning~\cite{warnecke2021machine}, preserving utility by
explicitly routing the forget signal into the retained
classifier heads.

\noindent\textbf{Refusal Vector (RV).}
RV computes the mean direction of forget-group image embeddings
and projects it out of all image embeddings at inference time,
making the forget group's visual features invisible to the
classifier:
\begin{equation}
    \tilde{\phi}(x) =
    \mathrm{normalize}\!\left(
        \phi(x) - \left(\phi(x) \cdot \mathbf{v}\right)\mathbf{v}
    \right),
    \label{eq:rv}
\end{equation}
where $\phi(x) = \mathrm{enc}_\mathrm{img}(x) /
\|\mathrm{enc}_\mathrm{img}(x)\|$ and
$\mathbf{v} = \mathrm{normalize}(\boldsymbol{\mu}_f -
\boldsymbol{\mu}_r)$ is the unit vector pointing from the mean
retain embedding $\boldsymbol{\mu}_r$ toward the mean forget
embedding $\boldsymbol{\mu}_f$.
The same projection is applied to $\mathbf{W}$ for consistency.
This method is directly inspired by concept erasure in
representation space~\cite{belrose2023leace} and aligns with
the notion of refusal vectors in large language models \cite{arditi2024refusal}.
In CLIP's embedding space,
$\cos(\boldsymbol{\mu}_f, \boldsymbol{\mu}_r) = 0.929$,
indicating that the forget and retain mean embeddings are nearly collinear. As established by Belrose~\etal~\cite{belrose2023leace}, perfect linear erasure requires orthogonality between these directions a condition not satisfied here explaining why RV achieves only partial forgetting ($\text{FA} = 64.30\%$) and why increasing projection strength beyond $\lambda{=}1.0$ paradoxically restores the original geometry, as reported in \S\ref{sec:analysis}.
\section{Experiments}
\label{sec:experiments}

\subsection{Experimental Setup}

\noindent\textbf{Dataset.}
We use CelebA~\cite{liu2015deep}, a large-scale face attribute dataset
with 202,599 images and 40 binary attributes.
We construct four intersectional demographic groups from the
\texttt{Young} and \texttt{Male} binary attributes:
\textit{Young Female}~(YF), \textit{Young Male}~(YM),
\textit{Old Female}~(OF), and \textit{Old Male}~(OM).
All evaluations are conducted on the official CelebA test split
(19,962 samples) reported in Table~\ref{tab:dataset_stats}.

\begin{table}[h]
\centering
\caption{CelebA demographic group statistics (test split).}
\label{tab:dataset_stats}
\begin{tabular}{lcc}
\toprule
\textbf{Group} & \textbf{Label} & \textbf{Count} \\
\midrule
Young Female & YF & 10,331 \\
Young Male   & YM &  4,783 \\
Old Female   & OF &  1,916 \\
Old Male     & OM &  2,932 \\
\midrule
Total        &    & 19,962 \\
\bottomrule
\end{tabular}
\end{table}

\noindent\textbf{Models.}
We evaluate three CLIP variants~\cite{radford2021learning} in the
zero-shot classification setting: ViT-B/32 ($d{=}512$),
ViT-B/16 ($d{=}512$), and ViT-L/14 ($d{=}768$).
All models are used in their released form without fine-tuning.
For each model, we construct a four-way zero-shot classifier from
the text prompts \textit{``a photo of a young woman''},
\textit{``a photo of a young man''},
\textit{``a photo of an old woman''}, and
\textit{``a photo of an old man''}. These prompts are encoded into
a normalized weight matrix $\mathbf{W} \in \mathbb{R}^{4 \times d}$,
as described in \S\ref{sec:formulation}.

\noindent\textbf{Forget Set.}
We designate $\mathcal{D}_f$ as all test samples belonging to the
\textit{Young Female} group ($|\mathcal{D}_f| = 10{,}331$), the
largest demographic slice and the group on which all three models
achieve their highest baseline accuracy ($\geq 97\%$), indicating
consistent representational dominance across model scales.

\noindent\textbf{Baselines.}
We report each unmodified model as the reference point.
Since our unlearning methods are zero-shot and require no retraining,
a retrain baseline is not applicable. Instead, we compare three
zero-shot unlearning methods that differ in where and how they modify
the classifier: Prompt Erasure~(PE) and Prompt Reweighting~(PR)
operate at the text embedding level, while Refusal Vector~(RV)
operates at the image embedding level, as described in
\S\ref{sec:methods}.

\subsection{Main Results}

\begin{table*}[t]
\centering
\caption{Unlearning quality and bias redistribution across methods
and CLIP model variants.
FA = Forget Accuracy ($\downarrow$ better), RA = Retain Accuracy
($\uparrow$ better), DP = Demographic Parity Gap
($\downarrow$ better), RS = Redistribution Score ($\downarrow$ better) .
$\Delta$Acc columns show per-group accuracy shift (\%) relative to
the Original model; large non-zero values for YM/OF/OM indicate
bias redistribution.}
\label{tab:main_results}
\resizebox{\textwidth}{!}{%
\begin{tabular}{llcccccccc}
\toprule
\multirow{2}{*}{\textbf{Model}} &
\multirow{2}{*}{\textbf{Method}} &
\multirow{2}{*}{\textbf{FA (\%) $\downarrow$}} &
\multirow{2}{*}{\textbf{RA (\%) $\uparrow$}} &
\multirow{2}{*}{\textbf{DP $\downarrow$}} &
\multicolumn{4}{c}{\textbf{$\Delta$Acc per Group (\%)}} &
\multirow{2}{*}{\textbf{RS $\downarrow$}} \\
\cmidrule(lr){6-9}
& & & & & YF (forget) & YM & OF & OM & \\
\midrule
\multirow{4}{*}{ViT-B/32}
  & Original           & 98.76 & 54.54 & 0.7298
                       &    &    &    &    &  \\
  & Prompt Erasure     &  0.00 & 78.57 & 0.9697
                       & $-$98.76 & $+$0.88 & $+$71.19 & $+$0.03 & 24.03 \\
  & Prompt Reweighting &  0.00 & 82.75 & 0.9577
                       & $-$98.76 & $-$14.11 & $+$69.99 & $+$28.75 & 37.62 \\
  & Refusal Vector     & 64.30 & 33.14 & 0.5335
                       & $-$34.46 & $-$29.86 & $+$0.10 & $-$34.45 & 21.47 \\
\midrule
\multirow{4}{*}{ViT-L/14}
  & Original           & 97.28 & 59.70 & 0.6075
                       &    &    &    &    &  \\
  & Prompt Erasure     &  0.00 & 80.27 & 0.9765
                       & $-$97.28 & $+$0.61 & $+$61.12 & $+$0.00 & 20.57 \\
  & Prompt Reweighting &  0.00 & 79.82 & 0.9614
                       & $-$97.28 & $-$18.98 & $+$59.60 & $+$19.75 & 32.78 \\
  & Refusal Vector     & 71.01 & 38.43 & 0.5208
                       & $-$26.27 & $-$38.51 & $+$15.66 & $-$40.96 & 31.71 \\
\midrule
\multirow{4}{*}{ViT-B/16}
  & Original           & 97.22 & 57.08 & 0.6690
                       &    &    &    &    &  \\
  & Prompt Erasure     &  0.00 & 79.26 & 0.9614
                       & $-$97.22 & $+$0.73 & $+$65.81 & $+$0.00 & 22.18 \\
  & Prompt Reweighting &  0.00 & 77.39 & 0.9008
                       & $-$97.22 & $-$25.90 & $+$59.76 & $+$27.08 & 37.58 \\
  & Refusal Vector     & 68.35 & 36.52 & 0.5191
                       & $-$28.87 & $-$28.37 & $+$4.49 & $-$37.79 & 23.55 \\
\bottomrule
\end{tabular}}
\end{table*}

Table~\ref{tab:main_results} summarizes unlearning quality and bias
redistribution across all three methods and all three CLIP variants.

\noindent\textbf{Forgetting.}
Both Prompt Erasure and Prompt Reweighting achieve perfect forgetting
($\text{FA} = 0.00\%$) across all three models, completely suppressing
each model's ability to classify Young Female samples.
Refusal Vector achieves only partial forgetting in all cases
(FA = 64.30\%, 71.01\%, and 68.35\% for ViT-B/32, ViT-L/14, and
ViT-B/16 respectively), which we attribute to the high cosine
similarity between the forget and retain mean embeddings
($\cos(\boldsymbol{\mu}_f, \boldsymbol{\mu}_r) = 0.929$)
the forget direction is deeply entangled with the retained embedding
space and cannot be fully isolated without collateral damage.

\noindent\textbf{Utility Preservation.}
Prompt Reweighting achieves the highest retain accuracy for ViT-B/32
($\text{RA} = 82.75\%$) and remains competitive across all variants,
outperforming Prompt Erasure by explicitly routing the forget
embedding's mass into the retained classifier heads.
Refusal Vector suffers the largest utility drop across all models
($\text{RA} \leq 38.43\%$) because projecting the forget direction
out of the image embedding space degrades representations for all
groups, not just the forget group.

\noindent\textbf{Bias Redistribution.}
All three methods produce substantial redistribution in at least one retained group across all model variants, confirming that forgetting is not neutral with respect to fairness. Most strikingly, both PE and PR transfer over 60\% points of Young Female accuracy onto Old Female across all three models
(PE: $+71.19\%$, $+61.12\%$, $+65.81\%$; PR: $+69.99\%$,
$+59.60\%$, $+59.76\%$ for ViT-B/32, ViT-L/14, ViT-B/16
respectively), while Young Male is largely unaffected by PE
($\leq +0.88\%$ across all models).
This asymmetry  redistribution flowing along gender rather than
age boundaries  is consistent across all three CLIP scales and
is analysed geometrically in \S\ref{sec:analysis}.
Refusal Vector produces the lowest redistribution score for ViT-B/32
($\text{RS} = 21.47$) but at the cost of incomplete forgetting and
severely degraded retain accuracy, exposing a fundamental
\emph{forget--fairness tradeoff} that holds across model scales.

\subsection{Bias Redistribution Visualization}

\begin{figure*}[t]
    \centering
    \includegraphics[width=\linewidth]{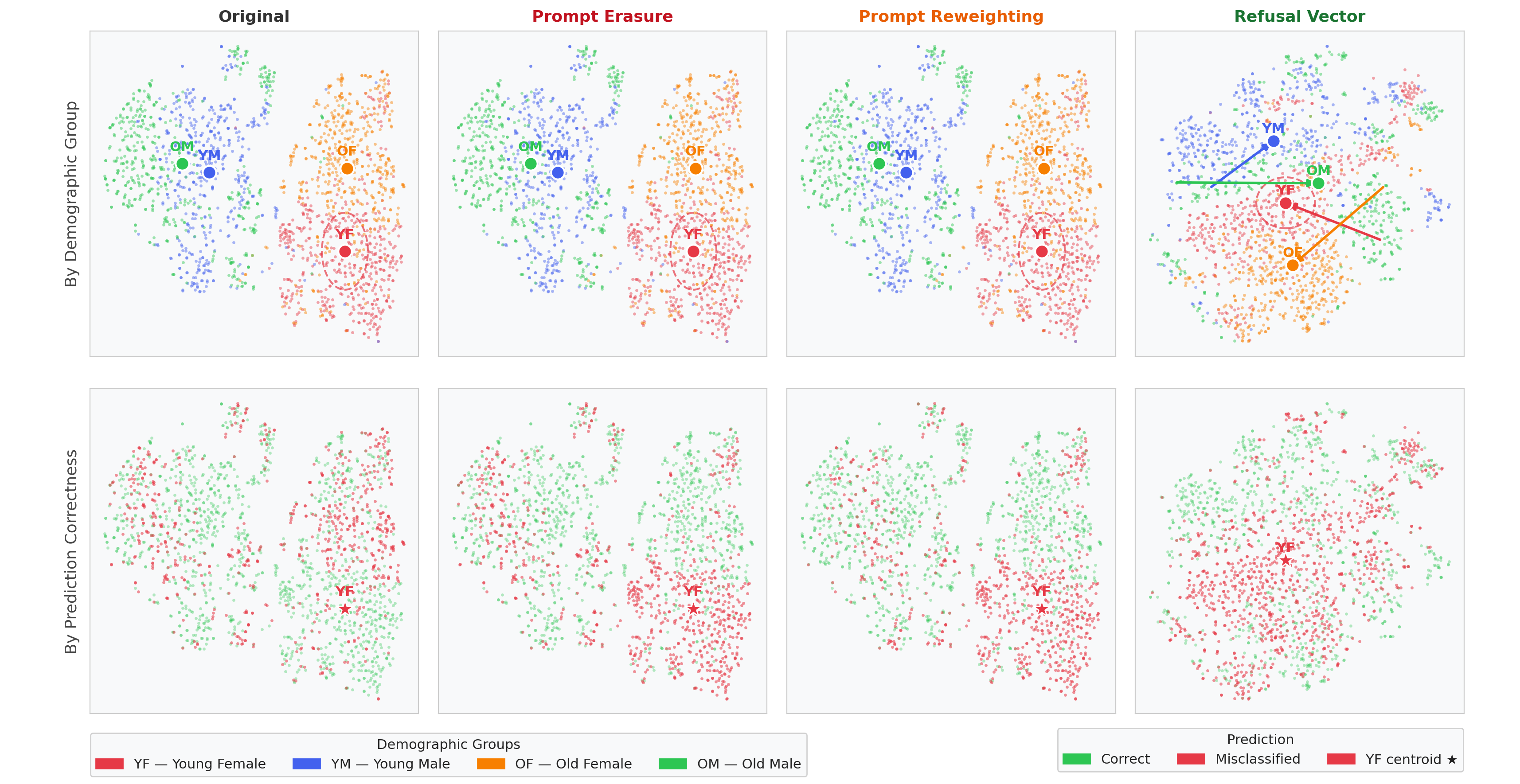}
    \caption{t-SNE \cite{van2008visualizing} projections of CLIP ViT-B/32 image embeddings
    (500 samples per group) before and after each method.
    Top row: colored by demographic group. Bottom row: colored by
    prediction correctness (green = correct, red = misclassified).
    Arrows indicate centroid drift relative to the Original.
    The proximity of YF and OF clusters explains why forgetting
    YF redistributes to OF rather than YM.}
    \label{fig:tsne}
\end{figure*}

Figure~\ref{fig:tsne} provides a geometric explanation for the
redistribution pattern observed in Table~\ref{tab:main_results}.
In the Original embedding space, the YF and OF clusters occupy
adjacent regions, while YM occupies a clearly separated region
despite sharing the same age axis as YF.
This geometry directly predicts the redistribution direction:
suppressing YF forces ambiguous images toward the nearest retained
cluster, which is OF rather than YM.
After Prompt Erasure and Prompt Reweighting, the correctness panels
(bottom row) confirm this  red misclassified points concentrate
in the YF region and shift toward OF, not YM.
Refusal Vector partially disrupts this geometry but at the cost
of collateral damage across all groups.

\begin{figure}[t]
    \centering
    \includegraphics[width=\linewidth]{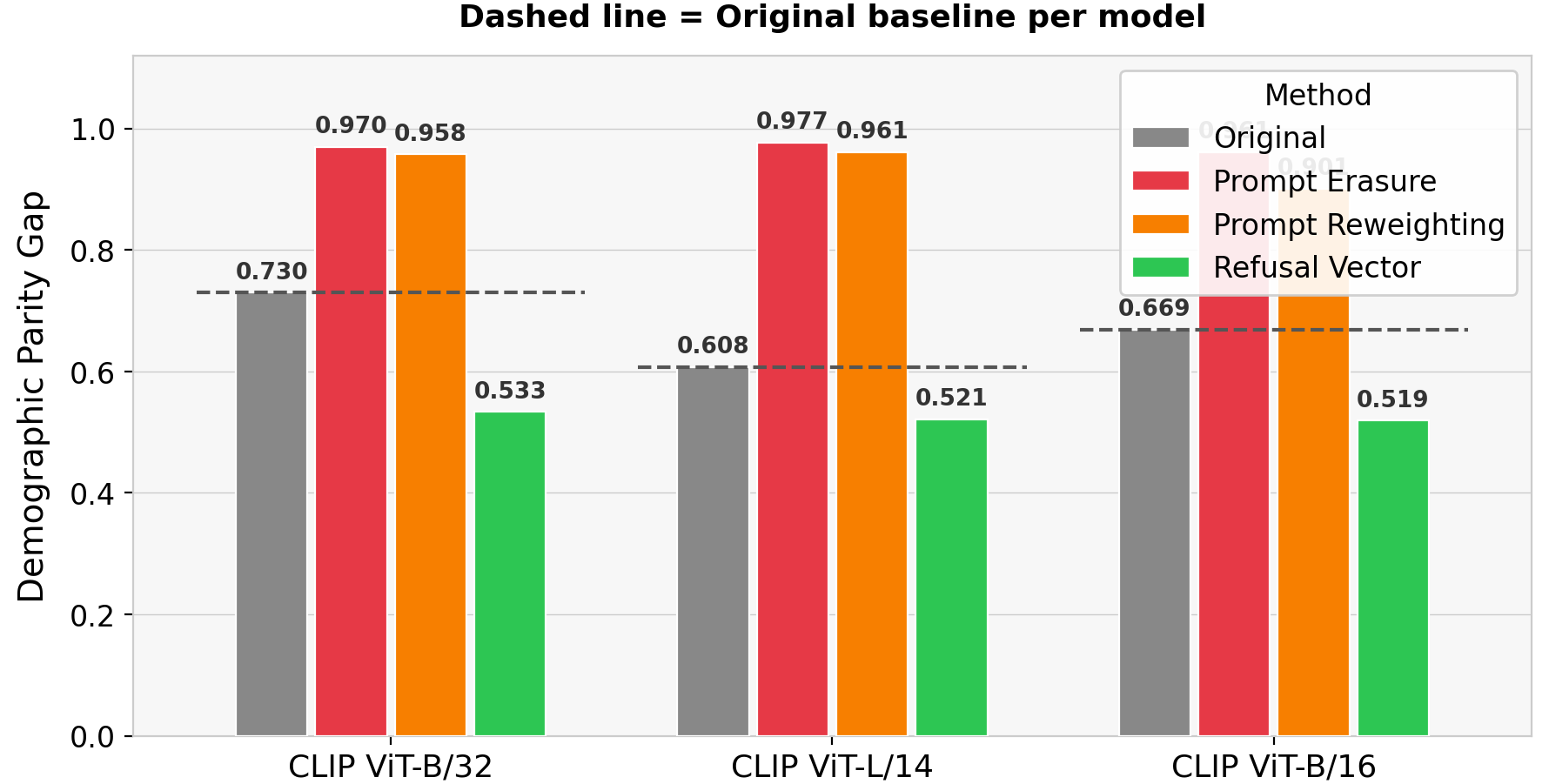}
    \caption{Demographic Parity Gap (DP) before and after unlearning
    for each method across all three CLIP variants. Prompt Erasure
    and Prompt Reweighting consistently worsen fairness despite
    perfect forgetting, while Refusal Vector is the only method
    that reduces the parity gap across all models
    (ViT-B/32: $0.730 \rightarrow 0.534$,
     ViT-L/14: $0.608 \rightarrow 0.521$,
     ViT-B/16: $0.669 \rightarrow 0.519$).}
    \label{fig:dp_gap}
\end{figure}

Beyond the geometric explanation, Figure~\ref{fig:dp_gap} shows
that bias redistribution incurs a fairness cost that is
consistent across all three CLIP model scales.
Prompt Erasure and Prompt Reweighting both substantially increase
the demographic parity gap despite achieving perfect forgetting,
because classification mass shifts to OF rather than being
distributed evenly.
Refusal Vector is the only method that reduces the parity gap, but
it does so at the cost of incomplete forgetting and the largest
utility drop among the three methods, confirming a fundamental
\emph{forget--fairness tradeoff}.

\begin{figure*}[t]
    \centering
    \includegraphics[width=0.92\linewidth]{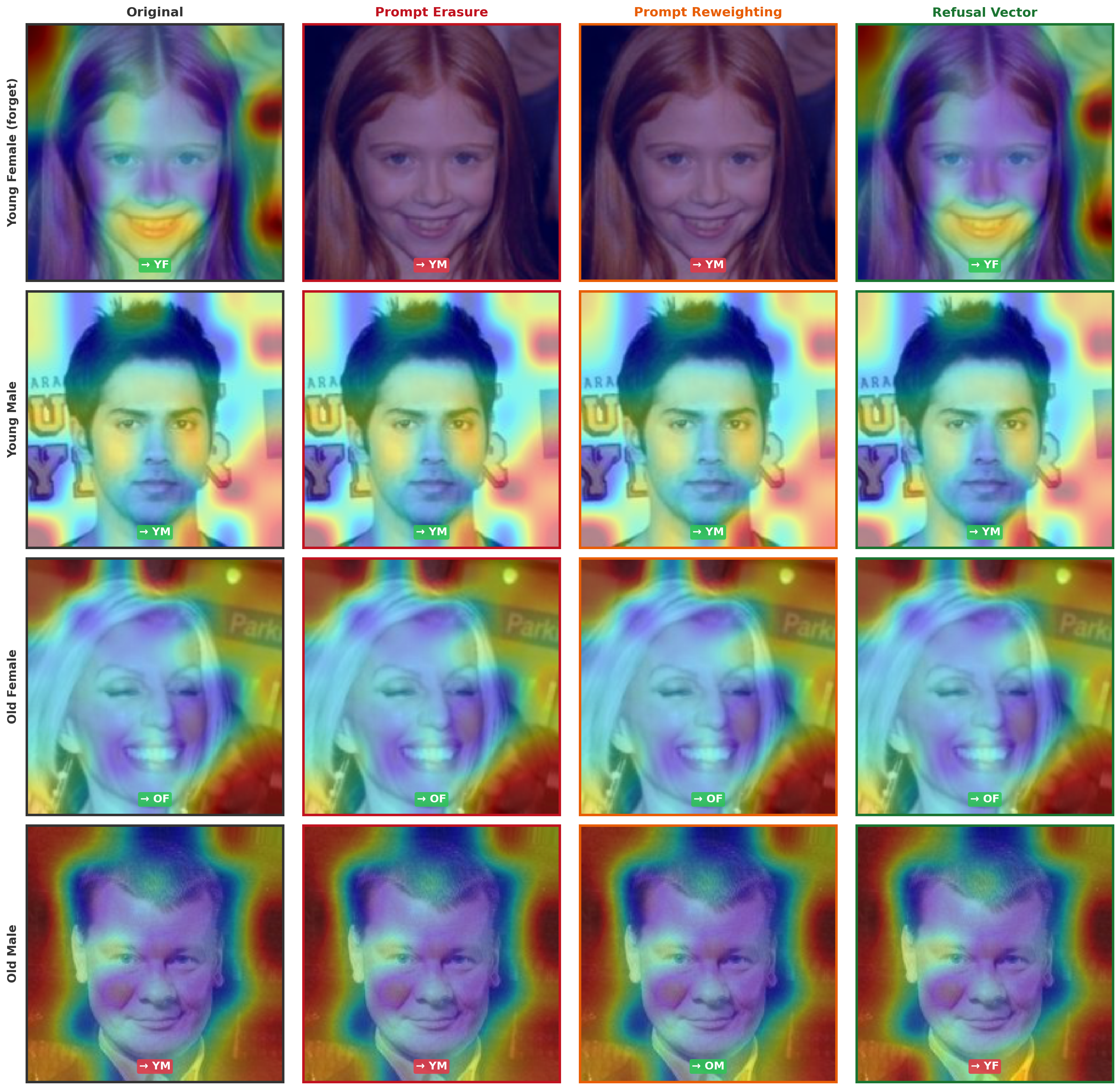}
    \caption{Patch similarity heatmaps for one representative face per demographic group across four conditions (Original, PE, PR, RV). Each heatmap shows cosine similarity between image patches and group text prompt embedding. The YF heatmap vanishes after PE and PR (no signal), while retained group heatmaps remain stable. Badge color indicates prediction correctness (green = correct, red = misclassified).
    }
    \label{fig:heatmaps}
\end{figure*}

Figure~\ref{fig:heatmaps} provides instance-level evidence of
redistribution through patch-level similarity heatmaps for one
representative face from each group.
For the YF face, the heatmap response is fully suppressed after PE
and PR, and the face is reassigned to OF, matching the
population-level redistribution observed in
Table~\ref{tab:main_results}.
For retained groups, the heatmaps remain largely stable under PE
and PR, suggesting that redistribution arises mainly at the
classifier rather than the feature level.
Under RV, the Old Male face is erroneously shifted toward the YF
direction, revealing collateral damage specific to
projection-based unlearning.

\subsection{Geometric Analysis of Redistribution}

\begin{figure}[t]
    \centering
    \includegraphics[width=0.9\linewidth]{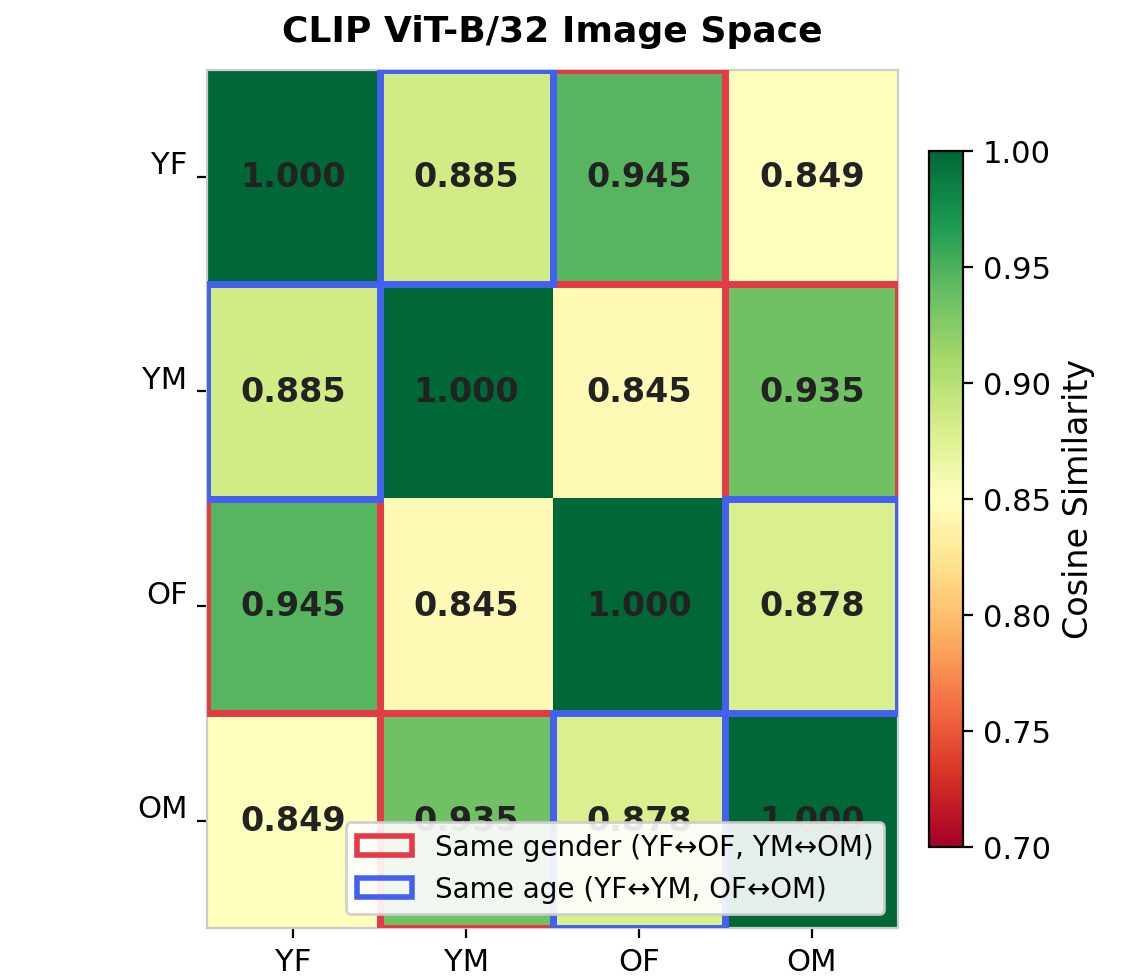}
    \caption{Pairwise cosine similarity between group mean image
    embeddings in CLIP ViT-B/32. Same-gender pairs
    (YF$\leftrightarrow$OF $= 0.945$,
    YM$\leftrightarrow$OM $= 0.935$, red borders) are consistently
    more similar than same-age pairs
    (YF$\leftrightarrow$YM $= 0.885$,
    OF$\leftrightarrow$OM $= 0.878$, blue borders), geometrically
    explaining why redistribution follows gender rather than
    age boundaries.}
    \label{fig:embedding_sim}
\end{figure}

\noindent\textbf{Embedding Space Geometry.}
To understand \emph{why} redistribution follows gender rather than
age boundaries, we compute pairwise cosine similarities between group
mean image embeddings in CLIP ViT-B/32's representation space
(Figure~\ref{fig:embedding_sim}).
The results reveal a clear gender-dominant geometry: same-gender
pairs (YF$\leftrightarrow$OF $= 0.945$,
YM$\leftrightarrow$OM $= 0.935$) are consistently more similar than
same-age pairs (YF$\leftrightarrow$YM $= 0.885$,
OF$\leftrightarrow$OM $= 0.878$).
This six-point gap directly predicts the redistribution direction 
when YF is suppressed, probability mass flows to the nearest retained
cluster, which is OF (same gender) rather than YM (same age).
This geometry is a property of CLIP's pretraining data, not of
the unlearning methods themselves.

\begin{figure}[t]
    \centering
    \includegraphics[width=0.92\linewidth]{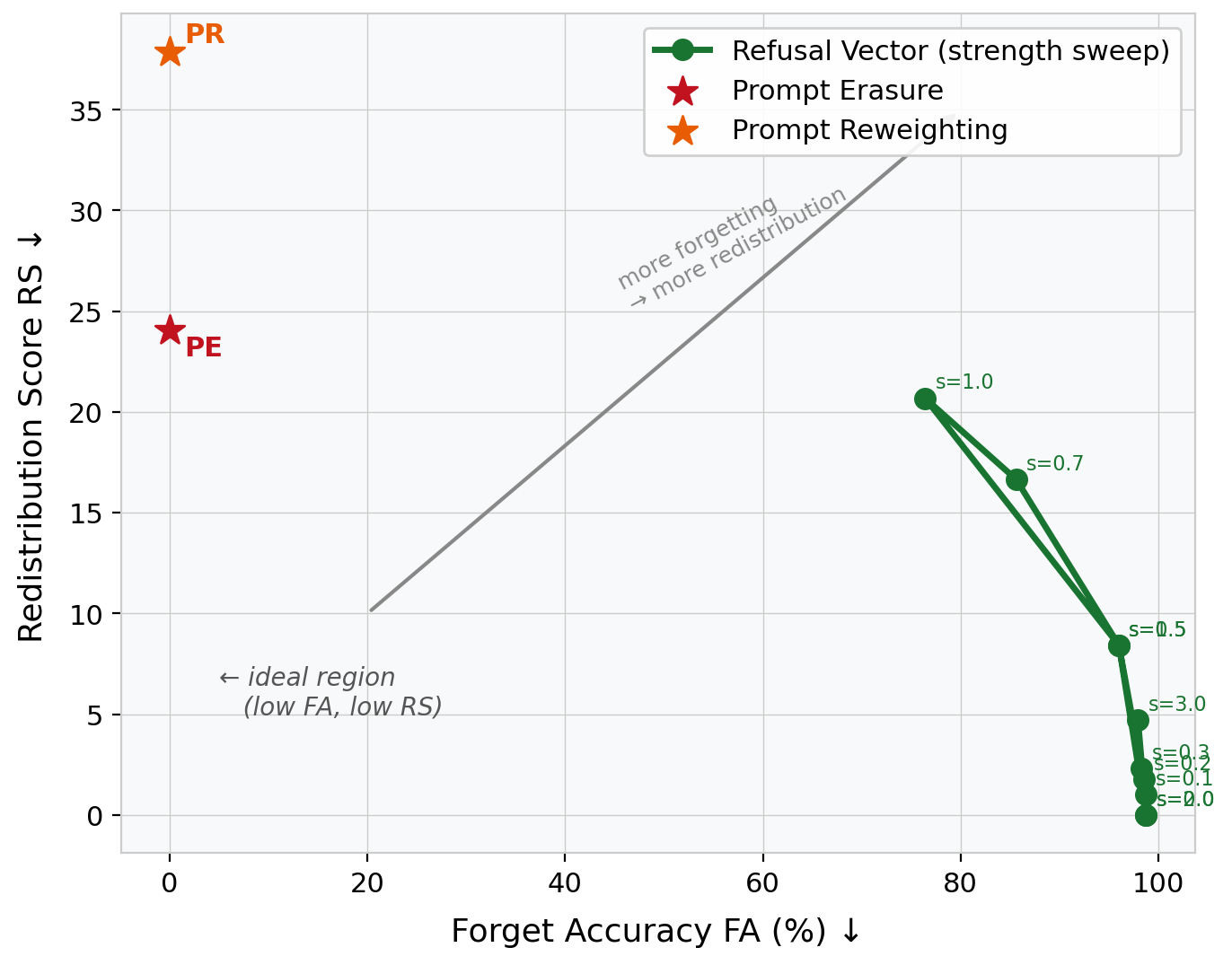}
    \caption{Forget--Fairness tradeoff curve. FA (x-axis,
    $\downarrow$ better) vs.\ RS (y-axis, $\downarrow$ better).
    The RV strength sweep traces a non-monotonic curve from no
    forgetting ($\lambda{=}0$) to peak forgetting ($\lambda{=}1.0$)
    and back due to over-projection at high strength. PE and PR are
    fixed points achieving perfect forgetting but high
    redistribution. The ideal operating point (lower-left) is
    unreachable by any zero-shot method evaluated here.}
    \label{fig:tradeoff}
\end{figure}

\noindent\textbf{Forget--Fairness Tradeoff Curve.}
Figure~\ref{fig:tradeoff} illustrates the tradeoff between Forget
Accuracy (FA) and Redistribution Score (RS) across all three
methods, where RV is additionally examined over projection
strengths $\lambda \in \{0.0, 0.1, 0.2, 0.3, 0.5, 0.7, 1.0, 1.5,
2.0, 3.0\}$.
PE and PR achieve perfect forgetting ($\text{FA} = 0\%$), but incur
substantial redistribution ($\text{RS} = 24.03$ and $37.62$,
respectively), placing them in the upper-left region of the
tradeoff space.
The RV sweep traces a non-monotonic curve: increasing the
projection strength initially improves forgetting while increasing
redistribution, whereas beyond $\lambda = 1.0$ the projection
overshoots, partially restoring the original geometry and reducing
both FA and RS.
No operating point achieves both low FA and low RS, indicating that
the two objectives are fundamentally in tension.
Overall, the tradeoff curve shows that none of the zero-shot
methods considered here resolves this tension without explicit
fairness constraints.

\section{Analysis and Discussion}
\label{sec:analysis}

\subsection{Which Method Redistributes Most?}

Prompt Reweighting produces the most severe redistribution
($\text{RS} = 37.62$) despite being the most conservative method:
it achieves the highest retain accuracy ($\text{RA} = 82.75\%$),
but amplifies imbalances across all retained groups
(OF: $+69.99$~pp, OM: $+28.75$~pp, YM: $-14.11$~pp), showing that
high utility does not preclude fairness degradation.
Prompt Erasure produces more concentrated redistribution
($\text{RS} = 24.03$): nearly all surplus mass flows to Old
Female ($+71.19$~pp), while Young Male and Old Male are virtually
unaffected, consistent with OF being the nearest retained centroid
to YF in CLIP's embedding space
(Figure~\ref{fig:embedding_sim}).
Refusal Vector achieves the lowest redistribution
($\text{RS} = 21.47$) and is the only method that improves
demographic parity ($0.7298 \rightarrow 0.5335$), but it does so
by degrading all groups uniformly rather than cleanly isolating the
forget group.

\subsection{Why Does Redistribution Follow Gender, Not Age?}

Both PE and PR transfer over 60 percentage points onto Old Female
across all three CLIP scales, while Young Male is nearly unaffected
($\leq +0.88$~pp)  the opposite of age-based intuition.
The pairwise similarity analysis (Figure~\ref{fig:embedding_sim})
explains why: same-gender pairs (YF$\leftrightarrow$OF $= 0.945$,
YM$\leftrightarrow$OM $= 0.935$) are more similar than
same-age pairs (YF$\leftrightarrow$YM $= 0.885$,
OF$\leftrightarrow$OM $= 0.878$), a six-point gap that directly
predicts the redistribution direction.
This gender-dominant geometry is consistent across ViT-B/32,
ViT-L/14, and ViT-B/16, confirming it is a property of CLIP's
pretraining data rather than a specific architecture.
Predicting where bias will go therefore requires auditing how
groups are arranged in the embedding space.

\subsection{The Geometric Impossibility of Perfect Erasure}

Refusal Vector does not achieve $\text{FA} = 0\%$, which is
consistent with the high alignment between the forget and retain
means, $\cos(\boldsymbol{\mu}_f, \boldsymbol{\mu}_r) = 0.929$.
As discussed by Belrose~\etal~\cite{belrose2023leace}, exact
linear erasure is most favorable when the forget and retain
directions are orthogonal, a condition that is not met here.
The RV strength sweep (Figure~\ref{fig:tradeoff}) is consistent
with this interpretation: beyond $\lambda = 1.0$, stronger
projection is associated with a partial restoration of the original
geometry, and $\text{FA} = 0\%$ is not observed.

\subsection{The Forget--Fairness Tradeoff}

More complete forgetting produces more severe redistribution across
all methods and model scales.
PE and PR achieve $\text{FA} = 0\%$ but worsen demographic parity
from $0.7298$ to $0.9697$ and $0.9577$ respectively.
RV reduces parity to $0.5335$ but leaves $\text{FA} = 64.30\%$.
No method simultaneously achieves low FA, high RA, and low RS 
the tradeoff is a geometric consequence of the pretrained embedding
space, not an artifact of method design.

\subsection{Practical Implications}

\begin{enumerate}
    \item \textbf{Always evaluate per-group accuracy.}
    Aggregate RA masks large within-group degradation  PR achieves
    the highest RA ($82.75\%$) yet the worst RS ($37.62$~pp).

    \item \textbf{Report RS alongside FA and RA.}
    A method achieving $\text{FA} \approx 0\%$ at the cost of large
    RS has not solved the fairness problem  it has relocated it.

    \item \textbf{Match method to use case.}
    When complete forgetting is legally required~\cite{voigt2017gdpr},
    PE is preferable to PR due to lower RS at equivalent FA.
    When fairness preservation is the primary goal, RV offers the
    best parity gap reduction.

\item \textbf{Audit embedding geometry before unlearning.}
High $\cos(\boldsymbol{\mu}_f, \boldsymbol{\mu}_r)$ signals that
no zero-shot method will achieve forgetting without fairness
collateral damage.

\end{enumerate}
\section{Conclusion}
\label{sec:conclusion}

We introduced \emph{bias redistribution}  the phenomenon whereby
machine unlearning does not neutralize a demographic group's
representation but relocates it along the geometric boundaries of
the pretrained embedding space.
Experiments on CelebA across three CLIP variants and three zero-shot
unlearning methods show that redistribution consistently follows
gender rather than age boundaries, explained by a gender-dominant
geometry ($\cos(\text{YF},\text{OF}){=}0.945 >
\cos(\text{YF},\text{YM}){=}0.885$), and that projection-based
unlearning cannot achieve perfect forgetting when forget and retain
embeddings are nearly collinear
($\cos(\boldsymbol{\mu}_f, \boldsymbol{\mu}_r) = 0.929$).
No method simultaneously achieves complete forgetting, utility
preservation, and fairness  confirming a fundamental
forget--fairness tradeoff rooted in embedding geometry.
We introduce the \emph{redistribution score} as a first-class
fairness metric for unlearning evaluation, and call for future work
on unlearning objectives that explicitly constrain intersectional
demographic parity in vision-language models.

\subsection{Limitations}
Our study has three main limitations.
First, we evaluate only zero-shot unlearning methods that
operate without gradient-based retraining whether
gradient-based approaches such as NegGrad+ and
SCRUB~\cite{kurmanji2023towards} exhibit similar
redistribution patterns remains an open question.
Second, all experiments use a single dataset (CelebA) with
one designated forget group (Young Female) future work
should evaluate redistribution across multiple datasets,
varying group imbalance ratios, and different forget targets,
including minority groups where redistribution dynamics
may differ substantially.
Third, our demographic attribute space is binary and coarse
(age: Young/Old, gender: Male/Female) extending to
continuous or intersectional attributes beyond two axes is
an important direction for future work.

\begin{CJK}{UTF8}{gbsn}
\section{Acknowledgment}
This work was supported by NewraLab (苏州拟界智能科技有限公司, Suzhou, China), an AI research and development startup founded by \texttt{Yunusa Haruna}. The authors gratefully acknowledge the support of the NewraLab team throughout this research. We thank \texttt{Haoyu Bian} for helpful feedback and for assisting in proofreading and polishing the manuscript.
\end{CJK}

{
    \small
    \bibliographystyle{ieeenat_fullname}
    \bibliography{main}
}


\end{document}